\pdfoutput=1

\documentclass[11pt]{article}

\usepackage{acl}
\usepackage[tikz]{bclogo}
\usepackage{lipsum}
\usepackage{times}
\usepackage{latexsym}

\usepackage[T1]{fontenc} 

\usepackage[utf8]{inputenc}

\usepackage{microtype}

\usepackage{booktabs}
\usepackage{graphicx} 
\usepackage{graphics}
\usepackage{amssymb}
\usepackage{amsmath}
\usepackage{pifont}
\usepackage{makecell}
\usepackage{resizegather}
\usepackage{etoolbox}
\usepackage{booktabs,makecell,tabularx} 
\usepackage{enumitem}
\usepackage{mathtools}
\usepackage{cleveref}
\usepackage{multirow}
\usepackage{booktabs,array}
\usepackage{amsmath, bm}
\usepackage{color, colortbl}
\usepackage{xcolor}
\usepackage{url}
\usepackage{hyperref}
\usepackage{arydshln}
\usepackage{booktabs}
\usepackage{multirow}

\usepackage[ruled,vlined]{algorithm2e}         
\newcolumntype{P}[1]{>{\centering\arraybackslash}p{#1}}

\let\oldnl\nl
\definecolor{Gray}{gray}{0.9}
\definecolor{LightCyan}{rgb}{0.88,1,1}
\newcommand{\nonl}{\renewcommand{\nl}{\let\nl\oldnl}}

\newcolumntype{H}{>{\setbox0=\hbox\bgroup}c<{\egroup}@{}}

\title{
PairReranker: Pairwise Reranking for Natural Language Generation}
 
\author{
Dongfu Jiang$^\ddagger$\thanks{$~$ The work was done when Dongfu Jiang was visiting USC as an intern, mentored by Bill Yuchen Lin.} \qquad Bill Yuchen Lin$^\dagger$ \qquad Xiang Ren$^\dagger$
\\
\texttt{dongfu@zju.edu.cn, yuchen.lin@usc.edu, xiangren@usc.edu} \\
$^\ddagger$ Zhejiang University \quad 
$^\dagger$ University of Southern California
\\ 
}

\begin{document}
\maketitle

\begin{abstract}

Pre-trained language models have been successful in natural language generation (NLG) tasks. While various decoding methods have been employed, they often produce suboptimal results. We first present an empirical analysis of three NLG tasks: summarization, machine translation, and constrained text generation. We found that selecting the best output from the results of multiple decoding methods can significantly improve performance. To further improve reranking for NLG tasks, we proposed a novel method, \textsc{PairReranker}, which uses a single encoder and a pairwise loss function to jointly encode a source input and a pair of candidates and compare them. Experiments on three NLG tasks demonstrated the effectiveness and flexibility of \textsc{PairReranker}, showing strong results, compared with previous baselines.
In addition, our \textsc{PairReranker} can generalize to significantly improve GPT-3 (text-davinci-003) results  (e.g., 24.55\% on CommonGen and 11.35\% on WMT18 zh-en), even though our rerankers are not trained with any GPT-3 candidates.  
\footnote{We will release our code and data at \url{https://inklab.usc.edu/PairReranker}.}
\end{abstract}

\begin{table*}[th!]
\begin{center}
    \scalebox{0.95}{
    \begin{tabular}{@{}c | c || c c c || c c || c@{}}
        \toprule
        \multicolumn{2}{c}{\textbf{Datasets} $\rightarrow$} & \multicolumn{3}{c}{CNN / DailyMail} & \multicolumn{2}{c}{CommonGen} & \multicolumn{1}{c}{WMT18(zh-en)} \\
        \toprule
         Decoding method $\downarrow$ & \# Candidates & R-1 & R-2 & R-L & BLEU & CIDEr & BLEU  \\
        \midrule
        Top beam & 1 & 44.22 & 21.48 & 41.21 & 14.62 & 15.48 & 19.20\\
        \midrule
         Beam search & 15$\rightarrow$ oracle & 51.06 & 27.74 & 48.05 & 23.62 & 24.56 & 25.84 \\
         Diverse beam search & 15$\rightarrow$ oracle& 54.30 & 30.02 & 51.33 & 23.58 & 23.61 & 26.08 \\
         Top-k sampling & 15$\rightarrow$ oracle & 52.32 & 27.74 & 49.19 & 21.66 & 21.44 & 19.60 \\
         Top-p sampling & 15$\rightarrow$ oracle & 53.51 & 28.83 & 50.40 & 22.78 & 22.82 & 20.84 \\
         \midrule
         All & 60$\rightarrow$ oracle& 57.70 & 33.75 & 57.42 & 30.04 & 29.89 & 34.51 \\
         Gain ($\frac{all-top}{top}\%$) & - & 30.5\% & 57.1\% & 39.3\% & 105.6\% & 93.0\% & 79.7\%\\
        \bottomrule
    \end{tabular}}
    \caption{\label{tab:main_oracle_gap} The empirical analysis of the oracle performance of decoding methods for three typical NLG tasks: summarization (CNNDM), constrained sentence generation (CommonGen), and machine translation (WMT18). Top beam is one of the most common methods of selecting output from beam searching results. However, the oracle selections in the top 15 candidates from four decoding methods are much better. If we use these four decoding methods jointly, the oracle performance gain is tremendous, as shown in the bottom row.
    The base models used here: PEAGUS for CNNDM, T5-Large for CommonGen, and Opus-MT for WMT. }
\end{center}
\end{table*}
\section{Introduction}
\label{sec:intro}
 



Pre-trained encoder-decoder language models (LMs) have demonstrated their effectiveness in various natural language generation (NLG) tasks. BART \cite{lewis2019bart} employs a denoising autoencoder architecture, in which the model is trained to reconstruct the original sentence from corruptions introduced during the training process. T5 \cite{t5} conducted a comprehensive study of alternative pre-training methods. Additionally, task-specific pre-trained models, such as PEGASUS \cite{zhang2020pegasus} for summarization tasks, have demonstrated impressive performance.


In order to effectively utilize encoder-decoder LMs for NLG tasks, it is crucial to employ an appropriate decoding method. Various decoding approaches, including beam search, diverse beam search~\cite{Vijayakumar2016DiverseBS}, top-k sampling, and top-p sampling~\cite{Holtzman2019TheCC}, have been widely employed during inference. These methods, which can be classified as either greedy algorithms or sampling-based approaches, often produce suboptimal results. Previous studies \cite{cohen2019empirical,meister2020if, Ravaut2022SummaRerankerAM} have found that the top beam search results are frequently inferior to the oracle selections from the candidates generated by these methods. The oracle performance can be significantly improved by selecting the best output from the results of multiple decoding methods.


In Table~\ref{tab:main_oracle_gap}, we present an empirical analysis of three typical NLG tasks: summarization (CNN/DM), machine translation (WMT18), and constrained text generation (CommonGen). For example, using the PEAGUS model~\cite{zhang2020pegasus} on the CNNDM dataset~\cite{Hermann2015TeachingMT}, we find that the Rouge-2 score for the top beam search generation is only 21.48. However, by selecting the best output from the top 15 beam search results based on the maximum Rouge-2 score, the upper-bound performance can be increased to 27.74. Further improvement can be achieved by combining the results from four different decoding methods and selecting the best output for evaluation, resulting in a performance of 33.75, a 57\% increase over the top beam search results. Similarly, the T5-large model on the CommonGen dataset~\cite{lin2019commongen} can achieve a 93\% gain in CIDEr score; the Opus-MT model~\cite{Tiedemann2020OPUSMTB} can achieve a 79.7\% gain in BLEU on the WMT18 (zh-en) translation task~\cite{Bojar2018FindingsOT}. These oracle performances highlight the importance of re-ranking candidate outputs in order to further improve the performance of LMs in NLG tasks.


The gap between oracle selections and top-ranked outputs is often attributed to the exposure bias caused by the teacher-forcing paradigm in most auto-regressive models. Re-ranking candidates after decoding is a simple yet effective way to mitigate this gap, and there have been several recent efforts in this direction. For instance, SimCLS \cite{Liu2021SimCLSAS} trains a re-ranker using a simple contrastive training objective, which encodes the source text and each candidate output using the same encoder and scores each candidate based on the cosine similarity between the embeddings. Another successful approach is SummaReranker (SR)~\cite{Ravaut2022SummaRerankerAM}, which uses mixture-of-experts training to improve the re-ranker for multiple metrics simultaneously. The encoding module in SR is based on a cross-encoder, which takes the concatenation of a source and a candidate target as a single input sequence and produces a score for the candidate. This cross-encoder design leverages the attention layers in the Transformer architecture~\cite{vaswani2017attentionia} and further enhances the ability to score candidates in the context of the source inputs.


Prior reranking approaches, such as SimCLS and SummaReranker, have primarily focused on individually scoring each candidate output based on a given input. However, this approach is limited as it does not directly learn to differentiate between candidate outputs. This is a particularly pertinent issue when the candidates for reranking are already ranked highly by some decoding methods, as the differences between them may be difficult to detect. Therefore, we argue that a dedicated pairwise comparison method is necessary to identify and correct subtle differences between candidates, thus allowing for improved reranking.


To this end, we propose a novel reranking method, \textsc{PairReranker}, the key idea of which is to jointly encode a source input and a pair of candidates (i.e., three items) with a single encoder and then use a pairwise loss function for learning to compare. Extensive experiments conducted on three NLG tasks (i.e., summarization, translation, and constrained sentence generation) demonstrate that \textsc{PairReranker} outperforms the baseline methods by a consistent margin and is also compatible with very large language models such as GPT-3~\cite{} (text-davinci-003).
\textsc{PairReranker} not only outperforms the previous SOTA method SummaReranker on the summarization task, but also shows great generalization performance in the other two NLG tasks, which are not evaluated previously. 
In addition, our \textsc{PairReranker} can be transferred to improve GPT-3 results by 24.55\% and 11.35\% for CommonGen and WMT18 (zh-en) respectively, even though our rerankers are not trained with any GPT-3 candidates.  

To sum up,  our proposed \textsc{PairReranker} method is a highly effective and flexible tool for reranking candidates in natural language generation tasks. Its use of a single encoder and pairwise loss function allows for improved performance and compatibility with larger language models. The strong results in multiple NLG tasks and the ability to transfer to GPT-3 further demonstrate the versatility and potential of \textsc{PairReranker} for use in a variety of natural language processing applications.

\section{Problem Formulation}
\label{sec:problem_formulation}



Auto-regressive language models are trained by maximizing the log-likelihood for each token
$p\left ( y_t|y_{<t}, x\right )$, where $x$ is the give source sequence and $y$ is the ground-truth output. The overall training objective is
\begin{equation}
\theta^* = \mathop{\arg\max}_{\theta} \mathop{\mathbb{E}}_{x, y} {\left (\sum_{i}{\log{p(y_i|y_{<i},x})} \right )}
\end{equation}

However, during the inference time, the model is supposed to predict the next token based on the tokens generated by itself, which causes the mismatch between the training and inference stage. 
\begin{equation}
o_i = \mathop{\arg\max}_{o_i}p(o_i|o_{<i},x)
\end{equation}
where $o_i$ is the $i$-th output token generated by the model and $o_{<i}$ is the output tokens before $o_i$. This mismatch is referred to as exposure bias.

To bridge the gap brought by the exposure bias, reranking-based methods aim to select the candidate of higher quality generated by decoding methods. 
Let $\mathbb{D}=\{(x_1, y_1),...,(x_n,y_n)\}$ be the set of source-target text pairs for a specific language generation task. For a specific pair of data $(x,y)$, let $C=\{c_{1}, ..., c_{m}\} \in \mathbb{C}$ be the corresponding generated candidates for source $x$. Thus the candidate of the highest quality with respect to the evaluation metric $\mu \in \mathbb{M}$ is 
\begin{equation}
c^*=\mathop{\arg\max}_{c_i\sim C}\mu{(c_i, y)}
\end{equation}

Without access to references,
a reranker $f_{\phi}$ should  give a higher score for $c^*$ than any other candidate $c_i$, where $\phi$ denotes the training parameters of the reranker $f$.
That is, 
{\normalsize{
\begin{equation}
\phi^* = \mathop{\arg\max}_{\phi} \mathop{\mathbb{E}}_{x,c_i\sim C}{\left( f_{\phi}(c^*,x)-\sum_i{f_{\phi}(c_i,x)} \right )}.
\end{equation}
}
}

\noindent
An ideal reranker is thus expected to further select a better candidate than the default choices from decoding algorithms. It is especially important when we want to merge results from multiple different decoding algorithms and select the best ones.
Various training paradigms have been proposed to maximize this expectation. We will present them as the baseline methods in Sec.~\ref{baseline methods}.




\section{Methods}
\label{sec:methods}
In this section, we formulate the methods of previous baselines as well as their limitations. Then we propose a pair-based reranker and demonstrate the improvements on these limitations.  
\subsection{Baseline methods}
\label{baseline methods}

\textbf{SimCLS}~\citet{Liu2021SimCLSAS} treat the reranking as a traditional learning-to-rank problem. They first encode a given source text $x$ and each generated candidate $c_i\in C\in\mathbb{C}$ with the same language encoder respectively (i.e., using the hidden state of the first token at the last layer;), denoted as $h(x)$ and $h(c_i)$.
Then, they take the cosine similarity between the source and candidate, $\tilde s_i=\cos{(h(x),h(c_i))}$, as the predicted score.
Let $\hat s =\cos{(h(x),h(y))}$ denote the output score for the reference $y$.  
Applying the marginal ranking loss, the overall training objective thus becomes:
\begin{equation}
\begin{split}
\phi^*_{\operatorname{SimCLS}} &= \mathop{\arg\max}_{\phi} 
\left (
\sum_j{\max\left (0, \tilde s_i - \hat s \right )}\right.\\
&\left. +\sum_i{\sum_{j>i}\max\left (0, \tilde s_j - \tilde s_i + \lambda_{i,j} \right)}
\right )   
\end{split}    
\end{equation}
where predicted scores $s_i$ are sorted according to metric score $\mu(c_i,y)$, and $\lambda_{i,j}=(j-i)\lambda$, where $\lambda$ is a hyper-parameter.

\textbf{SummaReranker}~\citet{Ravaut2022SummaRerankerAM} framed the problem as a binary classification task and train the reranker to tell the best candidates from the other candidates. 
Besides, they also add a mixture-of-expert (MoE) layer for jointly optimizing their reranker with multiple metrics. 
Let $\tilde s^*$ denote the predicted score of the best candidate $c^*$
\begin{equation}
    \phi^*_{\operatorname{SR}}=\mathop{\arg\max}_{\phi} \left ( \log (\tilde s^*) - \sum_{i, c_i\ne c*}\log (1-\tilde s_i) \right )
\end{equation}

However, none of these methods have ever given a direct comparison of the candidates. Since candidate groups are highly homogeneous, we believe direct comparisons via the attention mechanism are necessary.

\subsection{Pairwise Reranking}

Note that the candidates are  
highly homogeneous which makes it difficult for the model to learn their difference. Different from traditional document retrieval where the document corpus is rich and heterogeneous, the search space of our problem only contains dozens of generated candidates from a normal language model. Therefore, how to train a reranker that could capture the subtle nuance among the candidates is what we will focus on.

Similar to the baselines, our method also follows the paradigm of two-stage training. 
In contrast, our goal is to train a reranker $f$ with the parameter $\phi$ that correctly identifies the better one from a given \textit{pair} of candidates in a direct comparison manner. 
We want our reranker to learn to focus on the differences between two candidates and select the better one.
Here we give a formal description of our method. 

As stated in Sec.~\ref{sec:problem_formulation}, let $\mathbb{D}=\{(x_1, y_1),...,(x_n,y_n)\}$ be the set of source-target text pairs for a specific language generation task. 
Given a data point $(x,y)$, let $C=\{c_{1}, ..., c_{m}\}$ be a set of corresponding generated candidates for source $x$. 
For each candidate, there are some specific metrics $\mathbb{M}$ that indicate their quality (e.g., BLEU score). For each metric $\mu \in \mathbb{M}$, the score of candidate $c_i$ is denoted as $\mu(c_i, y)$.

For a metric $\mu$, our model is expected to output two scores: $s_{ij}^i, s_{ij}^j=f_{\phi}(s,c_i,c_j)$ respectively for a certain pair of candidates $(c_i, c_j)$. 
We use $s_{ij}=\sigma(s_{i,j}^i-s_{ij}^j)$ to denote the model's confidence of $c_i$ is better than $c_j$, where $\sigma$ is the sigmoid function. Then we consider the learning problem as a multi-task binary classification problem:
\begin{equation}
    \mathcal{L}^{\mu} = -z_{ij}^i \log{\sigma(s_{ij}^i)} - (1-z_{ij}^j)\log{\sigma(s_{ij}^j)} 
\end{equation}
where $z_{ij}^i=1, z_{ij}^j=0$ if $\mu(c_i, y) \ge \mu(c_j,y)$ and $z_{ij}^i=0, z_{ij}^j=1$ if $\mu(c_i, y) < \mu(c_j,y)$. 
For all the metrics in $\mathbb{M}$, we take the average as the final loss:
\begin{equation}
    \mathcal{L}=\frac{1}{|\mathbb{M}|} \sum_i^{|\mathbb{M}|} {\mathcal{L}^\mu}
\end{equation}

\subsection{\textsc{PairReranker} Architecture}
\label{sec:architecture}

To better capture the difference among the highly homogeneous candidate groups, we propose to conduct in-context attention among each candidate pair. Before concatenating each segment, we truncate each segment to specific lengths in case the total length after concatenating exceeds the model's capacity. 
After truncating, we add special tokens $\texttt{<source>}$, $\texttt{<candidate1>}$, and $\texttt{<candidate2>}$ to the start of corresponding segments. 
By separating with default separator token $\texttt{</s>}$, we get the concatenated input form: \texttt{``<s><source>$x$}\texttt{</s>}\texttt{<candidate1>$c_i$</s>} \texttt{<candidate2>}
 \texttt{$c_j$</s>''},
where $x$ is the text of a source input and $c_i$ and $c_j$ are the text of two output candidates.

We then feed them into the language model encoder and get the final hidden states. Among these hidden states, we use the embeddings of special tokens $\texttt{<source>}$, $\texttt{<candidate1>}$, and $\texttt{<candidate2>}$ as their representations. 

\begin{figure}[t] \centering
    \centering
    \includegraphics[scale=0.85]{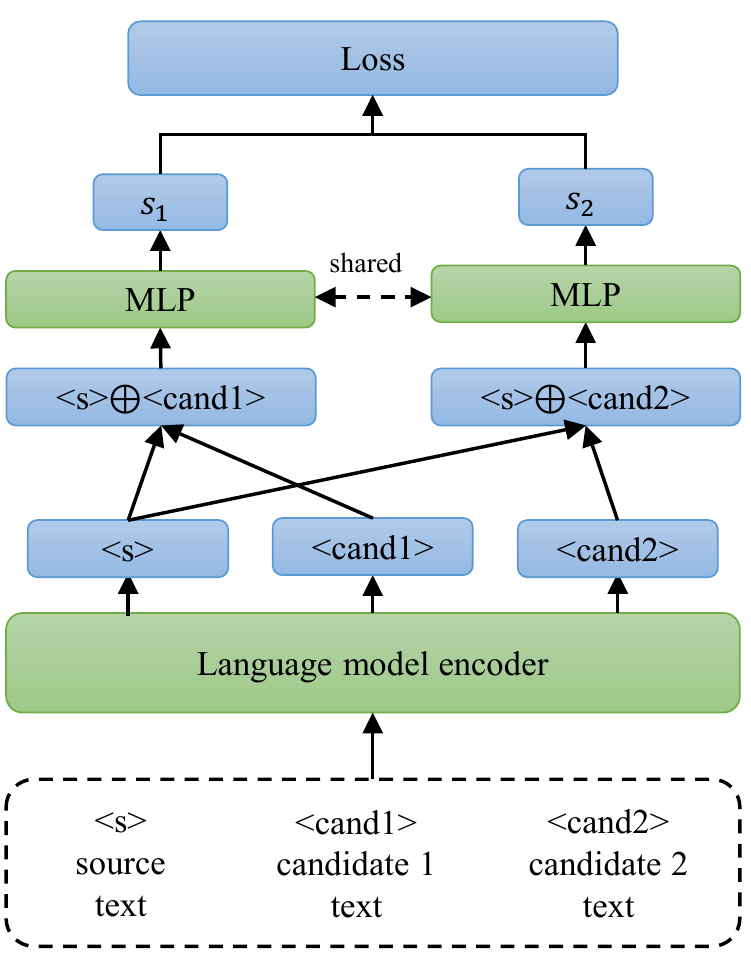}
    \caption{The encoding module of \textsc{PairReranker}.}
    \label{fig:modelarch}
\end{figure}

To compute the scores of the two candidates, we concatenate embeddings of $\texttt{<source>}$ with $\texttt{<candidate1>}$, and $\texttt{<source>}$ with $\texttt{<candidate2>}$ and feed them into a single-head layer that computes scores for each metric. 
Since there are $O(N^2)$ unique pair combinations, it is thus necessary for us to apply an effective sub-sampling strategy during both the training and the inference time for improving efficiency. 

During the training state, the selected pairs are supposed to be distinct enough to represent the characteristics of this small search space. Therefore, we select the best $k$ candidates and the worst $k$ candidates to form $k$ pairs for the training. In practice, we found that let $k=1$ is enough to get a decent result, which means for every source, we only use one pair of candidates for training. Besides, due to the position embeddings of the language model, the order of the candidates in a pair $(s,c_i,c_j)$ matters. That is, the comparison result of  $(s, c_i,c_j)$ and $(s,c_j,c_i)$ might not be consistent. 
Therefore, we also shuffle the order of candidates within each training pair so that the model learns to be consistent with itself.

In the inference stage, it would be too costly for us to compute the results of every possible pair and give the best candidate. 
Therefore, we instead do a single bubble run of comparisons to select the best candidate, which reduces the inference time complexity from $O(N^2)$ to $O(N)$. 
Let $k$ denote the best candidate index. For candidate group $C=\{c_1, \cdots, c_m\}$ after shuffling their positions, we initialize $k=1$. For each comparison with $c_i$, $i\in \{1,2,\cdots,m\}$, we compute the model's confidence $s_{ki}$ and $s_{ik}$. Then we update the best candidate index as follows:
\begin{equation}
    k = \left\{
        \begin{aligned}
            k&, & s_{ki} - s_{ik} > 0\\ 
            i&, & s_{ik} - s_{ki} > 0
        \end{aligned}
    \right.
\end{equation}
After doing $m-1$ times of comparison, we select $c_k$ as the best candidate.



The architecture of our model is illustrated in Figure~\ref{fig:modelarch}. Unlike the mixture-of-experts layer used in the work of \citet{Ravaut2022SummaRerankerAM}, we employ a 5-layer multi-layer perceptron (MLP) with the hyperbolic tangent activation function. The output dimension of the final layer is equal to the number of different metrics. It is worth noting that the weights of the two MLP layers are shared

\section{Evaluation}
\label{sec:eval}

\subsection{Tasks and data creation}

We evaluate the proposed method on the following public dataset: CNN/DailyMail, CommonGen, and WMT18. The statistics of these benchmarks are in Tab.~\ref{tab:dataset_stat}

\begin{table}[t]
\begin{center}
    \scalebox{0.75}{
    \begin{tabular}{c c c c   c c}
        \toprule
        \multirow{2}{3em}{Dataset} & \multicolumn{3}{c}{\# Examples} & \multicolumn{2}{c}{\# Words per example}\\
        ~ & Train & Val & Test & Source & Target \\
        \midrule
        CNN/DM & 287k & 13k & 11,490 & 856.56 & 70.05 \\
        CommonGen & 67k & 4k & 1,497 & 4.20 & 12.92 \\
        WMT18(zh-en) & 25m & 2k & 3,981 & 83.48 & 30.95 \\
        \bottomrule
    \end{tabular}}
    \caption{\label{tab:dataset_stat} Statistics of the three datasets}
\end{center}
\end{table}

\textbf{CNN/DailyMail} \citet{Hermann2015TeachingMT} is a dataset constructed from CNN and DailyMail websites. It is first used for machine-reading and comprehension, and later \citet{Nallapati2016AbstractiveTS} use it for abstractive summarization. Evaluation metrics are Rouge-1, Rouge-2, and Rouge-L. 
We use the public one from Hugging face Dataset\footnote{\url{https://huggingface.co/datasets/cnn_dailymail}}.

\textbf{CommonGen} \citet{lin2019commongen} is a dataset used for generative commonsense reasoning. It contains 79K commonsense descriptions where the language model is required to compose a realistically plausible sentence from given concepts. Evaluation metrics are BLEU-4, CIDEr and SPICE. Here we use the public one from Hugging Face Dataset\footnote{\url{https://huggingface.co/datasets/common_gen}}.

\textbf{WMT2018} \citet{Bojar2018FindingsOT} is a well-known dataset for evaluate machine translation. Here we use the Chinese-English split for evaluation. Evaluation metrics are BLEU. We use the public one from Hugging Face Dataset\footnote{\url{https://huggingface.co/datasets/wmt18}}.

\subsection{Base models}

For the summarization task on CNN/DailyMail dataset, we use the famous PEGASUS-large~\cite{zhang2020pegasus} and BART-large~\cite{lewis2019bart}, which have exhibited great ability for abstractive summarization. We use the public fine-tuned checkpoint from Hugging face\footnote{\url{https://huggingface.co/facebook/bart-large-cnn}}\footnote{\url{https://huggingface.co/google/pegasus-cnn_dailymail}}.

For the generative commonsense reasoning task on CommonGen dataset, we use the T5-large \cite{t5}. It's one of the vanilla baselines reported in \citet{lin2019commongen}. 

For the Chinese-English translation task on WMT2018 dataset, we use the public pre-trained opus-mt checkpoint\footnote{\url{https://huggingface.co/Helsinki-NLP/opus-mt-zh-en}}~\cite{Tiedemann2020OPUSMTB}.

\subsection{Evaluation setups}
\label{sec:evaluation_setup}

In this section, we talk about the training the testing paradigm of our reranker, including how we construct the training, validation, and testing dataset for our reranker, how we generate candidates, and what our experiment focuses on.
\begin{table}[t]
\begin{center}
    \scalebox{0.7}{
    \begin{tabular}{c c c c  c@{}}
        \toprule
         \textbf{Method} $\downarrow$ \textbf{Metric} $\rightarrow$ & R-1 & R-2 & R-L & Gain$_{R1}$   \\
        \midrule
        BART & 44.48 & 21.21 & 41.60 & - \\
        \textbf{PEGASUS} & 44.56 & 20.90 & 41.58 & - \\
        Gsum & 45.94 & 22.32 & 42.48 & -\\
        \midrule
        Gsum+RefSum & 46.18 & 22.36 & 42.91 & 1.18\%\\
        BART+SimCLS & 46.67 & 22.15 & 43.54 & 4.92\%\\
        PEGASUS+SummaReranker &47.16 & 22.55 & 43.87 & 5.83\%\\ 
        PEGASUS+\textbf{PairReranker}(Ours) & 47.29 & 22.77 & 44.06 & \textbf{6.12\%} \\
        \midrule
        \textbf{GPT-3} (text-davinci-003) & 37.96 & 15.51 & 34.39 & - \\
        \textbf{GPT-3}-oracle   & 45.46 & 22.83 & 42.04 & 19.76\% \\
        GPT-3+SummaReranker & 39.62 & 17.13 & 36.12 & 4.37\% \\
        GPT-3+\textbf{PairReranker}(Ours) & 40.41 & 17.44 & 36.79 & \textbf{6.45}\% \\ 
        \bottomrule
    \end{tabular}}
    \caption{\label{tab:cnndm_results} Model performance on \textbf{CNN/DailyMail}.}
\end{center}
\end{table}

To construct the training dataset for the reranker, we need to ensure the base model used to generate candidates on the training dataset should never have seen these candidates. Otherwise, the reranker will be trained on the candidates with higher quality compared to the candidates that it will be tested on, which we found will result in fairly bad performance. Therefore, following \citet{Ravaut2022SummaRerankerAM}, we first fine-tune the original non-finetuned pre-trained model on half of the training dataset, which gives us 2 half-finetuned base models that each of them has only seen their own half of the training dataset. Then we use them to generate candidates on their un-seen half of the training dataset using the decoding method talked about before. These generated candidates together form a whole training dataset with generated candidates that resemble the quality during the inference stage. 

During the inference stage, we directly adopt the public checkpoints that have been finetuned on the whole training dataset. We generate candidate candidates on the validation and testing dataset with this public checkpoint, which constitutes the validation and testing dataset on which our reranker does inference.
We use two decoding methods, beam search, and diverse beam search, in the experiments, following the prior work of SummaReranker. 
We generate 15 candidates for each decoding method for both training and inference. 

We train our reranker for 5 epochs. We use the Adafactor optimizer~\cite{Shazeer2018AdafactorAL}, with maximum learning rate being 1e-5. The warmup ratio is 5\% with a linear learning rate scheduler. Our training batch size is 64. The training finishes on a single RTX 8000 GPU in two days.

\subsection{Main results}
\label{main results}


\paragraph{Overall performance in summarization.}
Following the training and testing paradigm stated in section \ref{sec:evaluation_setup}, we briefly report the test results on the CNN/DM dataset in Tab.~\ref{tab:cnndm_results}. With fine-tuned PEGASUS-large as the base model. our method improves the candidates' quality by 6.12\% in Rouge-1, which is higher than our baseline SummaReranker. Besides, the performance gains in other metrics like Rouge-2 (8.94\%) and Rouge-L (5.96\%) are also obviously better.

\paragraph{Task generalization.}
In order to test the task generalization ability of our method, we here report the test results on CommonGen and WMT2018 (zh-en) in Tab.~\ref{tab:commongen_results} and Tab.~\ref{tab:wmt2018_results}. From the data in the table, our method also improves the candidates' quality significantly after reranking. We obtain a 2.90\% performance gain in CIDEr on the CommonGen dataset and a 3.87\% performance gain in BLEU on the WMT2018 dataset. 

We also report the performance of SummaReranker (our setup) on these two datasets. In contrast to the great performance on the summarization task, SummaReranker seems to fail to generalize well on other datasets. We also find that SummaReranker obtains a decreased gain on the CommonGen dataset (-1.23\% in CIDEr). The improvement on the translation dataset is not obvious (0.57\% in BLEU). We hypothesize that this is because of the average length of the candidates and the target text in these two datasets are all significantly smaller than the one in summarization (see in Tab.~\ref{tab:dataset_stat}). Therefore, the higher in-group similarity brought by the shorter length makes it harder for SummaReranker to capture their difference. On the contrary, our method with direct attention between a pair of candidates could easily tackle this problem. 
\begin{table}[t]
\begin{center}
    \scalebox{0.75}{
    \begin{tabular}{c c c c @{}}
        \toprule
          \textbf{Method} $\downarrow$ \textbf{Metric} $\rightarrow$  & BLEU & CIDEr  & Gain$_{\text{CIDEr}}$   \\
        \toprule
        T5-large & 14.62 & 15.48 &  - \\
        T5-large + SummaReranker & 14.13 & 15.29 & -1.23\% \\ 
        T5-large + \textbf{PairReranker} (Ours)
        & 15.30 & 15.93 & \textbf{2.90\%} \\
        \midrule
        GPT-3 (text-davinci-003) & 11.85 & 11.12 & - \\
        GPT-3 + oracle  & 20.34 & 19.26 & 73.20\% \\
        GPT-3 + SummaReranker & 13.71 & 13.21 & 18.79\% \\
        GPT-3 + \textbf{PairReranker} (Ours) & 14.39 & 13.85 & \textbf{24.55}\% \\ 
        \bottomrule
    \end{tabular}}
    \caption{\label{tab:commongen_results} Model performance on \textbf{CommonGen}. }
\end{center}
\end{table}

\paragraph{Transferring re-rankers to GPT-3.}

Besides, we also test our model's performance on the generated candidates from large language models like GPT-3. By directly applying the reranker that finishes training on the three datasets above to the GPT-3 generated candidates, we also report the performance gain in Tab.~\ref{tab:cnndm_results}, ~\ref{tab:commongen_results}, and~\ref{tab:wmt2018_results}.
We use OpenAI GPT-3 API to generate candidates by prepending a task-designed prompt to the source text. For all three datasets, we randomly sample 1000 data points from their test sets. Due to the cost as well as the length limitation of GPT-3 models, we prefer the data points with a shorter length. We run the generation using the text-davinci-003 model with the temperature set to 0.9. For each data point, we generate 15 candidates for reranking. The average quality of returned candidates and the oracle performance are reported in Tab.~\ref{tab:cnndm_results}, ~\ref{tab:commongen_results}, and~\ref{tab:wmt2018_results}.

We then directly run the inference on the GPT-3 data points using the rerankers trained above without adjustment. From the data reported in the table, we could see that the quality of the GPT-3 candidates is improved by a large margin compared to the average. Also, our method's performance gain is significantly larger than our baseline SummaReranker. For example, on the GPT-3 data points sampled from CNN/DM, our method obtain a gain of 6.45\%, whereas SummaReranker only obtains a gain of 4.37\%. And on the CommonGen's, our method obtains a gain of 24.55\% and SummaReranker only obtains a gain of 18.79\%.

\begin{table}[!t]
\begin{center}
    \scalebox{0.77}{
    \begin{tabular}{c c c c c}
        \toprule
          \textbf{Method} $\downarrow$ \textbf{Metric} $\rightarrow$  & BLEU & Gain     \\
        \midrule
        Opus-MT & 19.29 & - \\ 
        Opus-MT+SummaReranker & 19.40 & 0.57\% \\
        Opus-MT +\textbf{PairReranker (ours)} & 20.36 & 5.54\% \\ 
        \midrule
         GPT-3 (text-davinci-003) & 23.61 & - \\
        GPT-3 + oracle & 36.11 & 52.94\% \\
        GPT-3+SummaReranker & 25.08 & 6.22\% \\
        GPT-3+\textbf{PairReranker} & 26.29 & 11.35\% \\ 
        
        \hline
    \end{tabular}}
    \caption{\label{tab:wmt2018_results} Model performance on \textbf{WMT18 (zh-en)}.}
\end{center}
\end{table}
\paragraph{Abalation studies.}
Due to the order of the input format, changing the position of candidate 1 and candidate 2 might also change the results (Sec. \ref{sec:architecture}). In practice, we record the frequency of these 2 results being consistent and found that by simply shuffling the order of candidate 1 and candidate 2, our raranker could be consistent with itself more than 90\% of the time. Therefore, we believe our reranker is really learning to focus on the content of these two candidates.

Besides, our inference method, a single run of bubble sort with our reranker as the comparison function, also brings randomness to the final results. The initial order of the candidates might affect the final performance. In practice, we will shuffle the order of these candidates before the inference, which we believe will remove the bias of the initial order, which makes our final results more stable. 



\section{Related Work}
\label{sec:related}


How to exploit the encoder-decoder LMs for NLG tasks is one of the main focuses in natural language generation (NLG) research. Decoding methods like beam search and diverse beam search are widely used due to their famous ability of decoding candidates of high quality. However, previous studies have found that the top-beam candidate is usually not the best one.
Reranking methods have long been used to improve the quality of natural language generation.
\citet{Shen2004DiscriminativeRF} propose an effective ranking method for machine translation.
\citet{Kratzwald2018AdaptiveDR,Nogueira2019PassageRW} use passage reranking as the first stage of the question-answering system. \cite{Krishna2022RankGenIT} use reranking for Generative common sense generation task. 

For the summarization task, \citet{Liu2021RefSumRN} first provided a unified view of text summarization and summaries combination. After that, \citet{Liu2021SimCLSAS} proposed a contrastive learning framework to effectively rerank the summarization candidates. 
\citet{Ravaut2022SummaRerankerAM} then introduce multi-task learning in the reranking by applying a mixture-of-experts layer to the head of the language model encoder. 

As one of the most traditional and effective ranking methods, the pairwise ranking has shown its brilliant performance on a wide range of NLP tasks \cite{Jamieson2011ActiveRU}. Ranknet~\cite{Burges2005LearningTR} has been proposed as an effective method for a lot of ranking problems. The later improved LambdaRank~\cite{Burges2010FromRT} continues to show the great potential of pairwise ranking. However, though training in a pairwise way, they still compute the score for each item separately, which is why the pairwise method is outperformed by the listwise method. However, our method focuses on capturing the difference between a pair of data through the attention mechanism, which brings better performance.

Instead of focusing on reranker only, \citet{Liu2022BRIOBO} attempted to directly optimize the original language model by adding contrastive learning loss to the traditional cross-entropy loss as an adjustment to the fine-tuned language model. Following this, \citet{Shen2022JointGL} creatively trains the original language model and reranker jointly by applying the self-critic algorithm, which is one of the famous algorithms in reinforcement learning. That also points out a brand-new perspective on how to exploit the potential of the language model.
\section{Conclusion}
\label{sec:conclusion} 

Pre-trained encoder-decoder language models (LMs) have proven effective for natural language generation (NLG) tasks, but the performance of these models can be improved by using an appropriate decoding method during inference. Decoding approaches like beam search and top-k sampling often produce suboptimal results, and selecting the best output from the results of multiple decoding methods can significantly improve performance. Re-ranking candidate outputs after decoding can also help mitigate the gap between oracle selections and top-ranked outputs, and there have been several efforts in this direction, including SimCLS and SummaReranker. However, these approaches have limitations, including their inability to directly differentiate between candidates and their reliance on the quality of the initial rankings. 

In order to address these limitations, we propose a novel reranking method called \textsc{PairReranker}, which jointly encodes a source input and a pair of candidates with a single encoder and uses a pairwise loss function for learning to compare. Extensive experiments on three NLG tasks demonstrate that \textsc{PairReranker} outperforms baseline methods by a consistent margin.
Interestingly, trained rerankers can also be transferred and used with large language models such as GPT-3, which also produces a significant improvement. Overall, \textsc{PairReranker} is a highly effective and flexible method for reranking candidates in NLG tasks.



\bibliography{custom}
\bibliographystyle{acl_natbib}

\clearpage
\appendix
\begin{table*}[!h]
    \centering
    \scalebox{0.5}{
    \begin{tabular}{l l l}
        \hline
         \multicolumn{3}{c}{ CNN/Daily Mail}  \\
         \hline
        
\multicolumn{2}{l}{source(666)} &\makecell[l]{
(CNN)English actress Michelle Keegan has been named the sexiest woman in the world by British men's magazine FHM. \\
The 27-year-old actress is best known for her roles on the BBC series "Ordinary Lies" and the long-running British soap opera "Coronation Street." \\
Her "Coronation Street" character Tina McIntyre was the show's first character to act as a surrogate, according to IMDB. \\
Keegan is followed by reality television star and model Kendall Jenner, Oscar-winning actress Jennifer Lawrence,\\
Sports Illustrated swimsuit cover girl Kate Upton and British television host Caroline Flack, \\
who dated One Direction's Harry Styles for a few months. Sandra Bullock doesn't appear anywhere on FHM's list, \\
even though People magazine named her the "World's Most Beautiful Woman" a week ago. \\
The "FHM 100 Sexiest Women in the World" issue goes on sale Thursday, April 30.
} \\
\hline 

\multicolumn{2}{l}{target} &\makecell[l]{
British actress takes FHM's top spot in the list of 100 sexiest women in the world . \\
People's most beautiful woman is nowhere on the list .
} \\
\hline

\multirow{3}{8em}{Ours} & text & \makecell[l]{
"Ordinary Lies" actress Michelle Keegan is named sexiest woman in the world.<n>\\
Jennifer Lawrence, Kendall Jenner and Kate Upton also make FHM's list.<n>\\
Sandra Bullock doesn't appear on the list.
}\\
~ & scores & \makecell[l]{\text{Rouge-1: 44.07, Rouge-2: 17.54, Rouge-L: 40.68}} \\
~ & ranks & \makecell[l]{\text{Rouge-1: 1, Rouge-2: 1, Rouge-L: 1}} \\
\hline

\multirow{3}{8em}{SummaReranker} & text & \makecell[l]{
Michelle Keegan is best known for her roles on "Ordinary Lies" and "Coronation Street"<n>\\
Sandra Bullock doesn't appear on FHM's list.
}\\
~ & scores & \makecell[l]{\text{Rouge-1: 20.00, Rouge-2: 4.17, Rouge-L: 20.00}} \\
~ & ranks & \makecell[l]{\text{Rouge-1: 14, Rouge-2: 12, Rouge-L: 14}} \\
\hline

\multicolumn{2}{l}{source(156)} &\makecell[l]{
(CNN)Those poor fish must have been wondering what the heck was happening to them. \\
The Oregon Parks and Recreation Department has reported that a section of a fiberglass boat 20 or 30 feet long was \\
spotted off the state's coast this week and has been towed into harbor. \\
The debris is suspected to be from the earthquake and tsunami that hit Japan on March 11, 2011. \\
The boat fragment was found this week and towed to Newport, Oregon, where it is moored at a marina. \\
Inside were found -- more than four years and 4,000 miles later, if officials' suspicions are correct -- some specimens \\ 
of a variety of yellowtail jack fish normally found in Japanese waters. Biologists with the Oregon Coast Aquarium and \\
Oregon State University's Hatfield Marine Science Center inspected the debris while it was still at sea and determined \\
that the ecological threat posed by invasive species was small. The remnants of the boat will be dried out, \\
inspected further and taken to a landfill. But for the yellowtail jack fish, the journey is not over. \\
They'll be taken to the Oregon Coast Aquarium. 
} \\
\hline 

\multicolumn{2}{l}{target} &\makecell[l]{
Debris from boat to be dried, inspected and taken to landfill .\\
The debris contained fish normally found in Japanese waters .\\
The earthquake and tsunami hit Japan in March 2011 .
} \\
\hline

\multirow{3}{8em}{Ours} & text & \makecell[l]{
The debris is suspected to be from the 2011 earthquake and tsunami in Japan.<n>\\
The boat fragment was found this week and towed to Newport, Oregon.<n>\\
Inside were specimens of a variety of yellowtail jack fish normally found in Japanese waters.
}\\
~ & scores & \makecell[l]{\text{Rouge-1: 60.00, Rouge-2: 26.47, Rouge-L: 54.29}} \\
~ & ranks & \makecell[l]{\text{Rouge-1: 4, Rouge-2: 5, Rouge-L: 3}} \\
\hline

\multirow{3}{8em}{SummaReranker} & text & \makecell[l]{
A fiberglass boat fragment was found off the Oregon coast.<n>\\
It's suspected to be from the 2011 earthquake and tsunami in Japan.<n>\\
Inside were specimens of yellowtail jack fish.
}\\
~ & scores & \makecell[l]{\text{Rouge-1: 47.46, Rouge-2: 10.53, Rouge-L: 37.29}} \\
~ & ranks & \makecell[l]{\text{Rouge-1: 14, Rouge-2: 20, Rouge-L: 18}} \\
\hline

    \end{tabular}
} 

    \caption{Case studies.}
    \label{tab:cmp_ranking_case}
\end{table*}

\end{document}